\documentclass[11pt, letterpaper]{article}
\usepackage[utf8]{inputenc}
\usepackage[margin=1.5cm]{geometry}
\usepackage{titlesec}
\usepackage{tabu}
\usepackage{longtable}
\usepackage{enumitem}
\usepackage{amssymb}
\usepackage{xcolor}
\usepackage{graphicx}
\usepackage{amsmath}
\usepackage{amssymb}
\usepackage{biblatex}
\usepackage{tabularx}
\usepackage{xurl}
\usepackage{caption}
\usepackage{booktabs}
\usepackage{arabtex}
\usepackage{utf8}
\setcode{utf8}
\usepackage{ragged2e}

\newlist{selectlist}{itemize}{2}
\setlist[selectlist]{label=$\square$,leftmargin=*,noitemsep,topsep=0pt}

\newcommand{\mline}[1]{\begin{tabular}{@{}c@{}}#1\end{tabular}}

\usepackage{lmodern}

\usepackage{hyperref}
\hypersetup{
    colorlinks=true,
    linkcolor=blue,
    filecolor=magenta,      
    urlcolor=blue,
}
 
\titleformat{\section}[block]{\hspace{1em}\bfseries}{\thesection.}{0.5em}{} 
\titleformat{\subsection}[block]{\hspace{1em}}{\thesubsection}{0.5em}{}

\newif\iftherewashmode\therewashmodetrue
\newcommand\hmodeckeckautoref{\ifhmode\else\global\therewashmodefalse\fi\autoref}
\newcommand\sfcodefork{%
  \iftherewashmode\expandafter\@firstoftwo\else\global\therewashmodetrue\expandafter\@secondoftwo\fi{%
    \ifnum\spacefactor=1000 \expandafter\@firstoftwo\else\expandafter\@secondoftwo\fi
  }{\@secondoftwo}%
}%

\begin{document}

\hypertarget{target2}{}

{\huge\textbf{RGB Arabic Alphabets Sign Language Dataset}} \\

\vskip 0.5cm

\textbf{Authors}\\

\begin{itemize}
    \item Muhammad Al-Barham \textsuperscript{a,*}
    \item Adham Alsharkawi \textsuperscript{b}
    \item Musa Al-Yaman \textsuperscript{b}
    \item Mohammad Al-Fetyani \textsuperscript{c}
    \item Ashraf Elnagar \textsuperscript{d}
     \item  Ahmad Abu Sa’aleek \textsuperscript{e}
      \item Mohammad Al-Odat \textsuperscript{f}
\end{itemize}

\vskip 0.5cm
\textbf{Affiliations}\\ 
\begin{itemize}
    \item \textsuperscript{a} MLALP Research Group, University of Sharjah, United Arab Emirates
    \item \textsuperscript{b} Mechatronics Engineering Department, The University of Jordan
    \item \textsuperscript{c} AppsWave for Information Technology, Jordan
    \item \textsuperscript{d} Department of Computer Science, University of Sharjah, United Arab Emirates
    \item \textsuperscript{e} Al-Wefaq Control Systems, Doha, Qatar
    \item \textsuperscript{f} Student Guidance Department, The University of Jordan, Jordan
\end{itemize}

\vskip 0.5cm

 \textbf{Corresponding author’s email address and Twitter handle}

\begin{itemize}
    \item muhammadal-barham@ieee.org
    \item twitter: @MuhammadBarham\_
\end{itemize}

\vskip 0.5cm
\textbf{Keywords}\\ 

\vskip 0.5cm
\begin{itemize}
    \item[$\bullet$] Sign-Language
    \item[$\bullet$] Dataset
    \item[$\bullet$] Deaf
    \item[$\bullet$] Arabic
    \item[$\bullet$] Alphabet
\end{itemize}

\newpage

\vskip 0.5cm
 \textbf{Abstract}\\ 
 
 \vskip 0.5cm

\begin{justify}
This paper introduces the RGB Arabic Alphabet Sign Language (AASL) dataset. AASL comprises 7,857 raw and fully labelled RGB images of the Arabic sign language alphabets, which to our best knowledge is the first publicly available RGB dataset. The dataset is aimed to help those interested in developing real-life Arabic sign language classification models. AASL was collected from more than 200 participants and with different settings such as lighting, background, image orientation,
image size, and image resolution. Experts in the field supervised, validated and filtered the collected images to ensure a high-quality dataset. AASL is made available to the public on Kaggle.\footnote{\url{https://kaggle.com/datasets/59761a7132888de252ded8443ced1c7fb21ad28be5598f1f6ca43c663c32b40b}}

\end{justify}

\vskip 0.5cm
\textbf{Specifications table}\\
\vskip 0.2cm 

\begin{longtable}{|p{40mm}|p{124mm}|}
\hline
\textbf{Subject}                & 
Computer Science, Computer Vision, Pattern Recognition
                               
                               \\

\hline                         
\textbf{Specific subject area}  & 
RGB-Image Based Arabic Sign Language Classification

\\
\hline
\textbf{Type of data}           &

                 Images
\\                                   
\hline
\textbf{How the data were acquired} &

Images in this dataset were acquired using different types of cameras (webcam, digital camera, and camera phone).

\\
\hline                         
\textbf{Data format}            &

Labelled filtered RGB images with different extensions ('.jpg': 6545, '.jpeg': 1211, '.JPG': 80, ,'.JPEG': 21)

\\                                                    
\hline
\textbf{Description of data collection}  &

Participants were asked to submit their captured images through a form. Arabic sign language alphabets are grouped into five main categories and each category consists of a number of Arabic sign language alphabets. Gestures of the Arabic sign language alphabets are shown to the participants to follow. The quality and suitability of submitted images are checked manually.

\vskip 0.2cm
\\                         
\hline                         
\textbf{Data source location}& 
          
Jordan.\\

\hline                         
\hypertarget{target1}
{\textbf{Data accessibility}}   &


The Data is available on Kaggle under CC BY-NC-SA 4.0,
publicly available via the link
\url{https://kaggle.com/datasets/59761a7132888de252ded8443ced1c7fb21ad28be5598f1f6ca43c663c32b40b}\newline

Data identification number: 
It will be provided once the paper is accepted and the dataset become publicly available. \newline

\vskip 0.2cm
                         
                         \\                         
\hline                         

\end{longtable}


\newpage

\textbf{Value of the Data}\\

\begin{justify}
\begin{itemize}
\itemsep=0pt
\parsep=0pt

\item[$\bullet$] The data is versatile as it is collected with different settings such as lighting, background, image orientation, image size, and image resolution.

\item[$\bullet$] The dataset is suitable for developing machine learning algorithms for Arabic sign language classification.

\item[$\bullet$] The dataset is verified and validated by experts in the field.

\item[$\bullet$] This dataset is - to our best knowledge - the first RGB high-resolution and publicly available dataset for Arabic sign language.


\end{itemize}
\end{justify}

\vskip 0.5cm

\textbf{Data Description}  \newline

\begin{figure}[htb!]
    \centering
    \includegraphics[width=\textwidth]{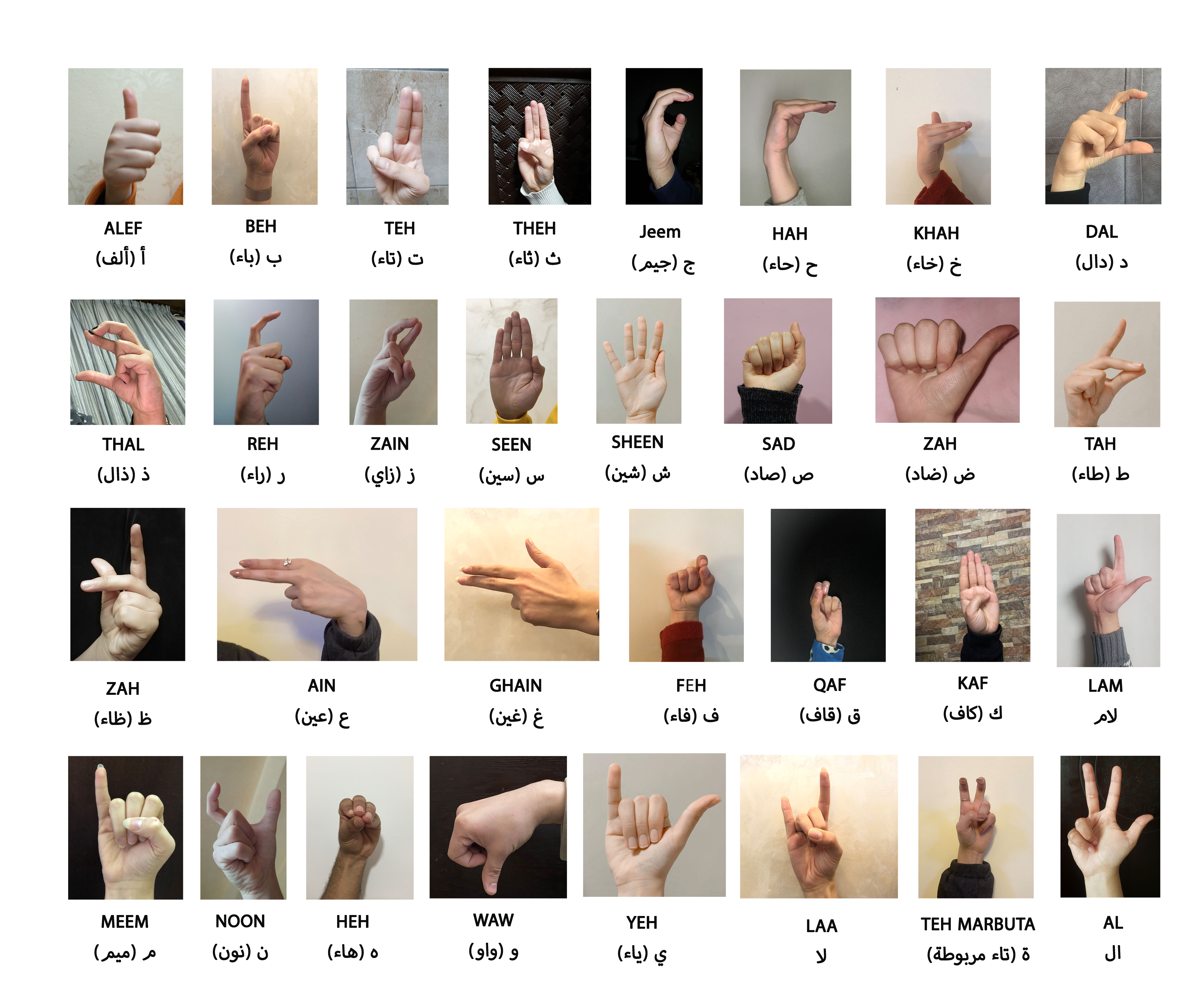}
    \caption{Sample from the dataset.}
    \label{fig:my_label}
\end{figure}

\begin{justify}
The RGB Arabic Alphabet Sign Language (AASL) dataset is the result of a collaborative effort among more than 200 participants who shared one or more alphabets. Most of the images were taken by different types of cameras including webcams, digital cameras, and phone cameras. The AASL dataset introduces 7,857 labeled images for the Arabic sign language. A group of Arabic sign language experts supervised, validated and filtered the images to ensure a high-quality dataset. \newline

The dataset is organized into 31 folders, each folder represents a single alphabet. Table~\ref{Tab:Table_1} highlights the number of images in each folder, while Fig~\ref{fig:my_label} presents a sample of images for different alphabets.
\end{justify}

\begin{table}[ht]
\caption{Dataset distribution.}

\begin{tabular}{|*4l|l|*4l|}

 \cmidrule{1-4}\cmidrule{6-9}
 \# & \mline{Letter name in \\English Script} & \mline{Letter name in \\Arabic Script} & \# of Images & & \# & \mline{Letter name in \\English Script} & \mline{Letter name in \\Arabic Script} & \# of Images \\ 
 
 \cmidrule{1-4}\cmidrule{6-9}
 
 1 & ALEF & \<أ  (ألف)> &
  287 & &
 17 & ZAH   & \<ض (ظاء)>&
 232  \\ 
 2 & BEH   & \<ب (باء)>&
 307 & &
 
 18 & AIN & \<ع (عين)>
 & 244 \\
 
 3 & TEH  & \<ت (تاء)>&
 226 & &
 
 19 & GHAIN & \<غ (غين)> 
 & 231 \\
 
 4 & THEH & \<ث (ثاء)>
  & 305 & &
  
 20 & FEH & \<ف (فاء)>
 & 255 \\
 
 5 & JEEM  & \<ج (جيم)>
 & 210 & &
 
 21 & QAF  & \<ق (قاف)>
 & 219 \\
 
 6 & HAH & \<ح (حاء)>
 & 246 & &
 
 22 & KAF  & \<ك (كاف)>
 & 264 \\
 
 7 & KHAH & \<خ (خاء)>
 & 250 & &
 
 23 & LAM & \<ل (لام)>
 & 260 \\
 8 & DAL & \<د (دال)>
 & 235 & &
 
 24 & MEEM & \<م (ميم)>
 & 253 \\
 
 9  & THAL & \<ذ (ذال)>
 & 202 & &
 
 25 & NOON & \<ن (نون)>
  & 237 \\
  
 10 & REH & \<ر (راء)> 
  & 227 & &
  
 26 & HEH & \<ه (هاء)>
  & 253 \\
 
 11  & ZAIN & \<ز (زاي)>
 & 201 & &
 
 27 & WAW  & \<و (واو)>
 & 249 \\
 
 12 & SEEN & \<س (سين)>
 & 266 & &
 
 28 & YEH & \<ي (ياء)>
 & 272 \\
 
 13 & SHEEN &\<ش (شين)>
 & 278 & &
 
 29 & TEH MARBUTA & \<ة (تاء مربوطة)>
 & 257 \\
 
 14 & SAD & \<ص (صاد)>
 & 270 & &
  
 30 & AL & \<ال>
 & 276 \\
 
 15 & DAD & \<ض (ضاد)>
 &  266 & &
 
 31 & LAA & \<لا> 
 & 268 \\
 
 16 & TAH & \<ط (طاء)>
 & 227 & &
 &  &  & \\ 
 
 \cmidrule{1-4}\cmidrule{6-9}
\end{tabular}
\label{Tab:Table_1}
\end{table}%

\newpage

\noindent
\vskip 0.5cm

\textbf{Experimental design, materials and methods:}
\newline
\begin{justify}

With the aim of contributing to the Arabic sign language classification, we asked experts in the field of ArSL interpretation to provide and verify ground-truth images that represent static ArSL alphabets. The experts also helped in providing tips on how to perform each of the alphabets.  

\end{justify}

\begin{justify}

An online form with a set of instructions was prepared for data collection. The alphabets were distributed into five different categories for the participants, the first 4 categories have 6 alphabets and the fifth and last category has the remaining 7 alphabets. Participants had the option to submit images of the alphabets that they felt comfortable performing them. Hence, there was not any restriction on the number of images that a participant should submit.  \\

The link to the online form was posted on different social media platforms. We had participants from schools and universities with different ages and genders. Images were captured by the participants using different types of cameras, backgrounds, light conditions, and image sizes. The identity of the participants was kept anonymous.

\end{justify}

\begin{figure}[htb!]
 \centering
 \small
\begin{tabular}{ccc}

\includegraphics[scale=0.7]{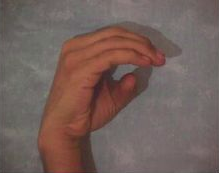}&
\includegraphics[scale=0.2]{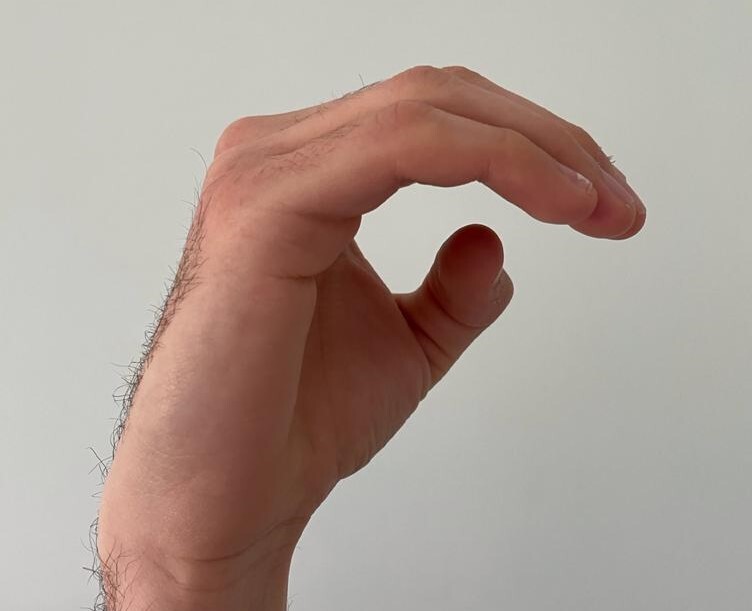} &
\includegraphics[scale=0.03]{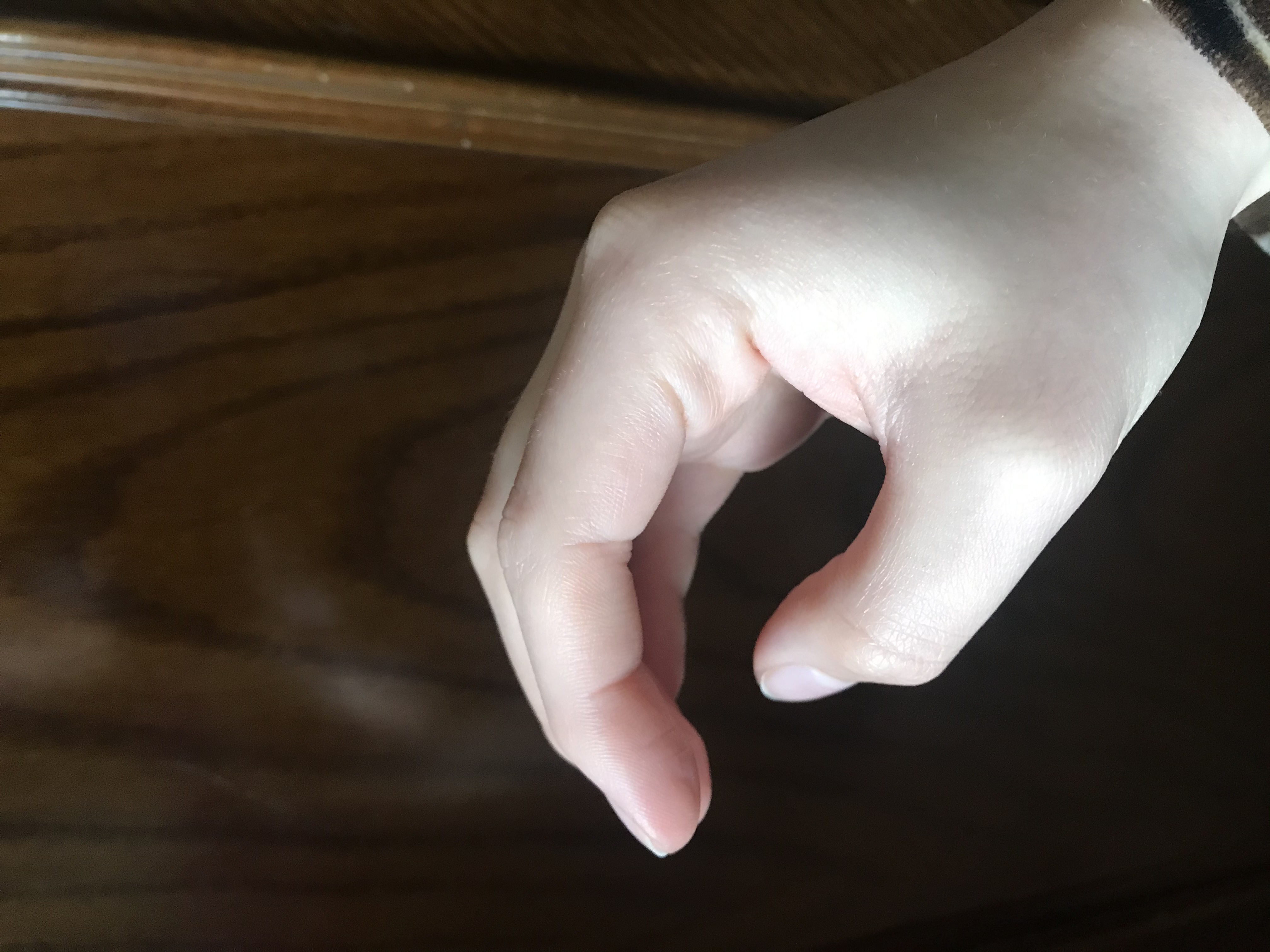}  \\

(1) Ground-Truth image. & (2) Correct image. & (3) Wrong image. \\

\end{tabular}

\caption{Geem ArSL alphabet.}
\label{Fig:Geem_alphabet}
\end{figure}

\begin{justify}

The data collection started in March 2022 and lasted for five months. Two of our research team were given the task of evaluating each and every submitted image manually. They were mainly responsible for checking the label of an image and the match between a submitted image and the ground-truth image of a particular alphabet. Fig~\ref{Fig:Geem_alphabet} shows an example of a ground-truth image of an alphabet (left), a correctly performed alphabet (center), and a wrongly performed alphabet (right).\\

The whole dataset then went through one final round of evaluation where one of our research team double-checked all submitted images for correctness. The evaluation process resulted in a dataset size reduction going from 8,042 images to 7,857 correct images.\\ 

Finally the whole dataset was labelled automatically by running a simple script. Each of the images is labeled as "AphabetName\_ID". The ID started from 0 till reaching the total number of images of a certain alphabet in a specific folder.\\

On a final note, images of our dataset are raw in nature, and thus interested researchers are left to perform any necessary processing they may need. Also, this work has been inspired by ArASL (Arabic Alphabets Sign Language) Dataset \cite{LATIF2019103777}.\\

\end{justify}






\noindent
\textbf{CRediT author statement}\\
\noindent

\vskip0.5cm


\noindent

\textbf{Muhammad Al-Barham}: Conceptualization, Validation, Methodology, Writing- Original draft preparation, Software, Data Curation\newline
\textbf{Adham Alsharkawi}: Writing- Reviewing and Editing\newline
\textbf{Musa Al-Yaman}: Conceptualization,  Writing- Original draft preparation, Resources\newline
\textbf{Mohammad Al-Fetyani}: Writing- Reviewing and Editing, Software, Data Curation \newline
\textbf{Ashraf Elnagar}: Writing- Reviewing and Editing\\
\textbf{Ahmad Abu Sa’Aleek}: Conceptualization, Methodology, Validation\\
\textbf{Mohammad Al-Odat}: Validation, Methodology\\

\noindent
\vskip0.5cm

\textbf{Acknowledgments}\\

\vskip0.3cm

\begin{justify}

We would like to thank the Student Counseling Department at the University of Jordan for their guidance on how to get the right and correct images based on their experiences. We would like also to thank Jana M. AlNatour and Raneem F. Abdelraheem for their help in the data collection process. \newline
\end{justify}



\printbibliography

@article{LATIF2019103777,
title = {ArASL: Arabic Alphabets Sign Language Dataset},
journal = {Data in Brief},
volume = {23},
pages = {103777},
year = {2019},
issn = {2352-3409},
doi = {https://doi.org/10.1016/j.dib.2019.103777},
url = {https://www.sciencedirect.com/science/article/pii/S2352340919301283},
author = {Ghazanfar Latif and Nazeeruddin Mohammad and Jaafar Alghazo and Roaa AlKhalaf and Rawan AlKhalaf}
}

\end{document}